\documentclass[sigconf]{acmart}

\AtBeginDocument{%
  \providecommand\BibTeX{{%
    \normalfont B\kern-0.5em{\scshape i\kern-0.25em b}\kern-0.8em\TeX}}}

\copyrightyear{2021}
\acmYear{2021}
\setcopyright{acmcopyright}
\acmConference[MM '21]{Proceedings of the 29th ACM International Conference on Multimedia}{October 20--24, 2021}{Virtual Event, China}
\acmBooktitle{Proceedings of the 29th ACM International Conference on Multimedia (MM '21), October 20--24, 2021, Virtual Event, China}
\acmPrice{15.00}
\acmDOI{10.1145/3474085.xxxxxxx}
\acmISBN{978-1-4503-8651-7/21/10}

\usepackage{multirow}
\begin{document}
\fancyhead{}
\title{HANet: Hierarchical Alignment Networks for Video-Text Retrieval}

\author{Peng Wu$^{1}$*,\quad Xiangteng He$^{2\dagger}$,\quad Mingqian Tang$^{2}$,\quad Yiliang Lv$^{2}$, \quad Jing Liu$^{1\dagger}$}

\makeatletter
\def\authornotetext#1{
\if@ACM@anonymous\else
    \g@addto@macro\@authornotes{
    \stepcounter{footnote}\footnotetext{#1}}
\fi}
\makeatother
\authornotetext{Work done during an internship at Alibaba Group.}
\authornotetext{Corresponding authors.}

\affiliation{
 \institution{\textsuperscript{\rm 1}Guangzhou Institute of Technology, Xidian University, Guangzhou, China}
 \institution{\textsuperscript{\rm 2}Alibaba Group, Hangzhou, China}
 \country{}
 }
 
\email{xdwupeng@gmail.com,neouma@163.com}
\email{{xiangteng.hxt, mingqian.tmq, yiliang.lyl}@alibaba-inc.com}

\def\authors{Peng Wu, Xiangteng He, Mingqian Tang, Yiliang Lv, and Jing Liu}

\renewcommand{\shortauthors}{Peng Wu, et al.}

\begin{abstract}
  Video-text retrieval is an important yet challenging task in vision-language understanding, which aims to learn a joint embedding space where related video and text instances are close to each other. Most current works simply measure the video-text similarity based on video-level and text-level embeddings. However, the neglect of more fine-grained or local information causes the problem of insufficient representation. Some works exploit the local details by disentangling sentences, but overlook the corresponding videos, causing the asymmetry of video-text representation. To address the above limitations, we propose a Hierarchical Alignment Network (HANet) to align different level representations for video-text matching. Specifically, we first decompose video and text into three semantic levels, namely event (video and text), action (motion and verb), and entity (appearance and noun). Based on these, we naturally construct hierarchical representations in the individual-local-global manner, where the individual level focuses on the alignment between frame and word, local level focuses on the alignment between video clip and textual context, and global level focuses on the alignment between the whole video and text. Different level alignments capture fine-to-coarse correlations between video and text, as well as take the advantage of the complementary information among three semantic levels. Besides, our HANet is also richly interpretable by explicitly learning key semantic concepts. Extensive experiments on two public datasets, namely MSR-VTT and VATEX, show the proposed HANet outperforms other state-of-the-art methods, which demonstrates the effectiveness of hierarchical representation and alignment. Our code is publicly available at \textcolor{blue}{\url{https://github.com/Roc-Ng/HANet}}.
\end{abstract}




\maketitle
\section{Introduction}
\begin{figure}[t]
  \centering
  \includegraphics[width=0.8\linewidth]{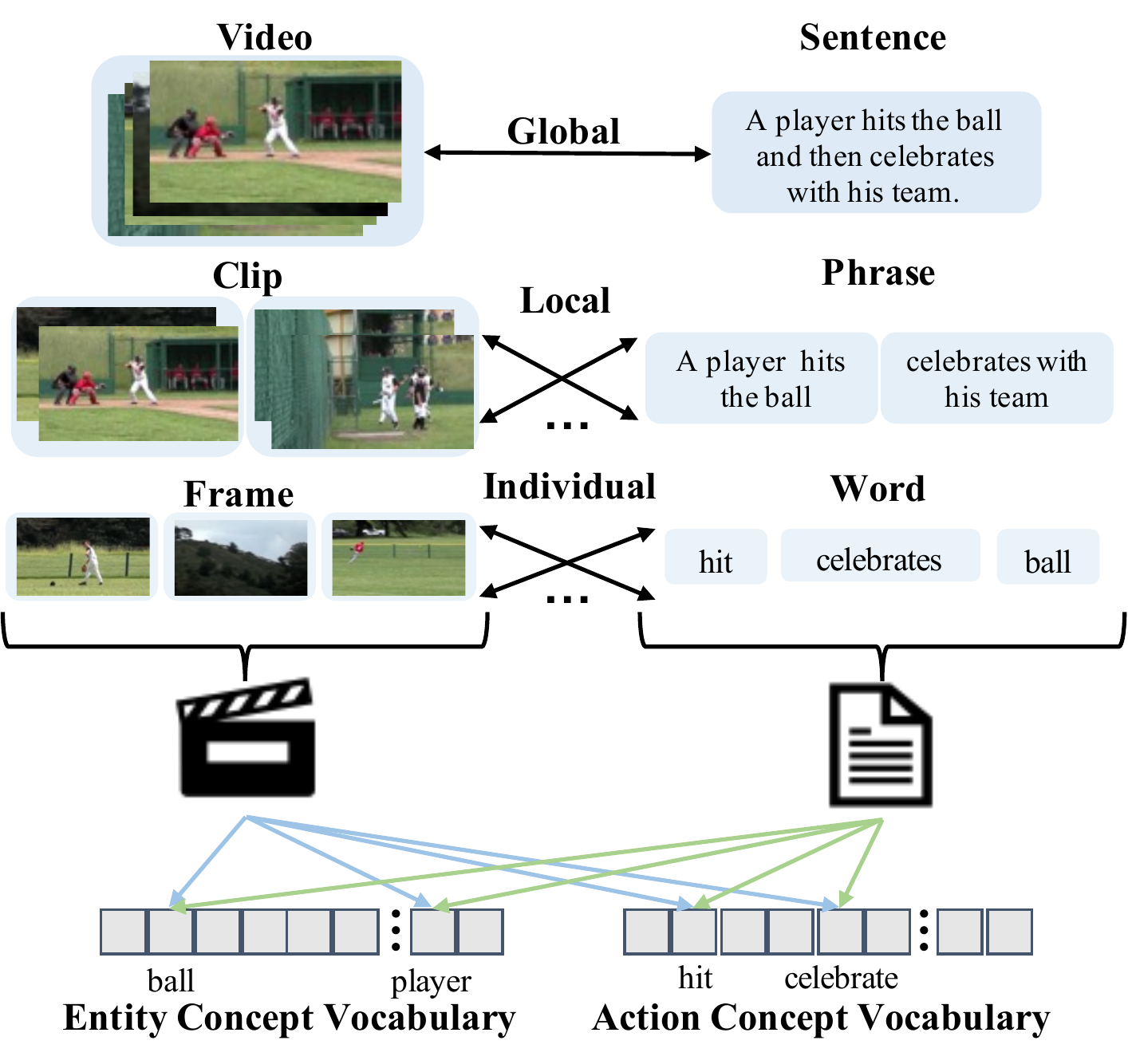}
  \caption{Illustration of the hierarchical alignment. The individual level focuses on the alignment between frame and word, local level focuses on the alignment between video clip and textual context,and global level focuses on the alignment between the whole video and text. 
  }
  \label{sample}
\end{figure}
“Hey, Siri, I want to watch a video of fighting in the desert.”
Recently, cross-modal retrieval has attracted increasing attention due to the explosive growth of online videos and advances in artificial intelligence technology. In addition to speech recognition, video-text retrieval is a key technique in the above scenario, which aims to search related videos given a natural-language sentence as the query. This task is challenging since the video and text are two different modalities, how 
to encode and match them in the joint space is the key.

Many efforts to make a reliable and accurate video-text retrieval system have been done. Recently, a typical practice is to encode videos and texts into compact representations and measure their similarities in a joint latent common space using metric learning. In this case, most existing works \cite{dong2019dual, mithun2018learning, li2019w2vv++, feng2020exploiting} focus on how to learn global representations of videos and texts, and achieve this goal by leverage various embedding networks, e.g., convolutional neural networks (CNN), gated recurrent units (GRU), Transformer, Bidirectional Encoder Representations from Transformers (BERT). However, such compact global representations neglect the more fine-grained or local information that existed in videos and texts, which may cause the problem of insufficient representation.

To mitigate this problem, some other works utilize local semantic information for fine-grained video-text retrieval. For example, Wray et al. \cite{wray2019fine} break the sentence into nouns and verbs using part-of-speech (pos) parsing and operate multiple cross-modal matching. Chen et al. \cite{chen2020fine} make further efforts for fine-grained retrieval by exploiting semantic alignments for both global event and local action and entity through hierarchical graph reasoning. Nevertheless, video-text retrieval is a cross-modal task, these methods only focus on text parsing, yet overlook video parsing, causing the asymmetry of video-text representation. Therefore, simultaneously parsing text and video is a more general solution.

To address the above issues, we propose the hierarchical alignment network (HANet), which aims to simultaneously parse text and video into different semantic levels, and then generates individual-, local- and global-level representations, finally hierarchically aligns different level representations in separate joint spaces. We illustrate the hierarchical alignment in Figure \ref{sample}.
The contributions of this paper are as follows:
\begin{itemize}
\item \textbf{Multi-semantic-level representation} is proposed to simultaneously parse videos and texts into different semantic levels, i.e. event-level on the whole video and text, action- and entity-level on the parsing parts of video and text, where we subtly parse videos by means of concept-based classification under weak supervision without other complicated steps. Here, the concept is from predefined concept vocabularies which drive the model to learn the detailed components in video and text, so that the cross-modal associations between video frames and keywords could be established via concepts, which makes our model more interpretable.
\item \textbf{Hierarchical Alignment} is proposed for cross-modal matching on top of multi-semantic-level representation. The individual level focuses on the alignment between frame and word on the concept-specific prediction features, local level focuses on the alignment between video clip and textual context, and global level focuses on the alignment between the whole video and text. Different level alignments not only capture fine-to-coarse correlations between videos and texts, but also take the advantage of the complementary information among three semantic levels.
\end{itemize}

We show the superiority of our HANet on two popular video-text retrieval datasets, i.e., MSR-VTT, VATEX. Without additional features and pre-training, HANet achieves clear performance improvements over state-of-the-art methods.

\section{Related Work}
\subsection{Video-Text Retrieval}
Video-text retrieval is a non-trivial branch of cross-modal retrieval \cite{song2019polysemous, he2019new, chen2019cross, xu2020proposal, patrick2020support, wray2021semantic, miech2020end, ging2020coot, lei2021less}. The typical  methods can be divided into three components, namely, text encoding, video encoding, and joint space learning. Recently, many works focus on designing powerful text and video encoding. For example, Dong et al. \cite{dong2019dual} proposed multi-level encodings of video and text in advance to learning shared representations. Li et al. \cite{li2019w2vv++} concatenated the bag-of-words vector, word2vec embedding, and Recurrent neural networks (RNN) vector as the final text representations. Similarly, Li et al. \cite{li2020sea} incorporated several sentence encoders and measured similarities in multiple encoder-specific common spaces rather than a single common space. Some works \cite{gabeur2020multi, liu2019use, miech2018learning} made full use of multimodal cues, e.g., motion, appearance, face, OCR, for video encoding. Other works attempted to decompose texts into some semantic phrases. Yang et al. \cite{yang2020tree} constructed a latent semantic tree to describe the text and used a temporal attentive encoder to obtain the temporal-attentive video representation. Xu et al. \cite{xu2015jointly} proposed a compositional semantics language model to parse the sentence into Subject-Verb-Object structure. Wray et al. \cite{wray2019fine} disentangled sentences into verbs and nouns for fine-grained video retrieval, and Chen et al. \cite{chen2020fine} disentangled texts into events, actions and entities. The above two methods are similar to ours, but they overlook disentangling videos and developing interpretability. Another two works \cite{liu2021hit, zhang2018cross} are also hierarchical models, however, the method in \cite{zhang2018cross} is not applicable to decompose single sentences, and the hierarchical transformer in \cite{liu2021hit} only focus on global features.

As for joint space learning, Chen et al. \cite{chen2020interclass} designed a new ranking loss that assigns weights to the relative similarities between positive and negative pairs. An analogous work \cite{wei2020universal} introduced a new polynomial loss with the universal weighting framework. With the help of these new losses, traditional methods achieve clear performance improvements. 

\subsection{Cross-modal Concept Learning}
In the last few years, cross-modal concept learning is usually utilized for a new challenge in  TRECVID, i.e., Ad-hoc Video Search (AVS). The majority of the top-ranked solutions \cite{le2016nii, ueki2017waseda_meisei, nguyen2017vireo, markatopoulou2017query} for AVS focused on computing the similarity between a textual query and a specific video via concepts. In terms of  video, constructing visual concept classifier to detect concepts; In terms of text, designing complex linguistic rules to extract relevant concepts. The merit of these concept-based AVS methods is the good interpretability, but the weakness is that using predefined concepts to describe videos and texts is insufficient. Recently, two hybrid works \cite{wu2020interpretable, dong2021dual} employed both concept-based and concept-free strategies and achieved better performance. Similar to them, Our HANet is also considered as a hybrid model. Another interesting work \cite{yu2017end} proposed a high-level concept word detector to generates a series of concept words as useful semantic priors for cross-modal tasks.

\begin{figure*}[t]
  \centering
  \includegraphics[width=12cm]{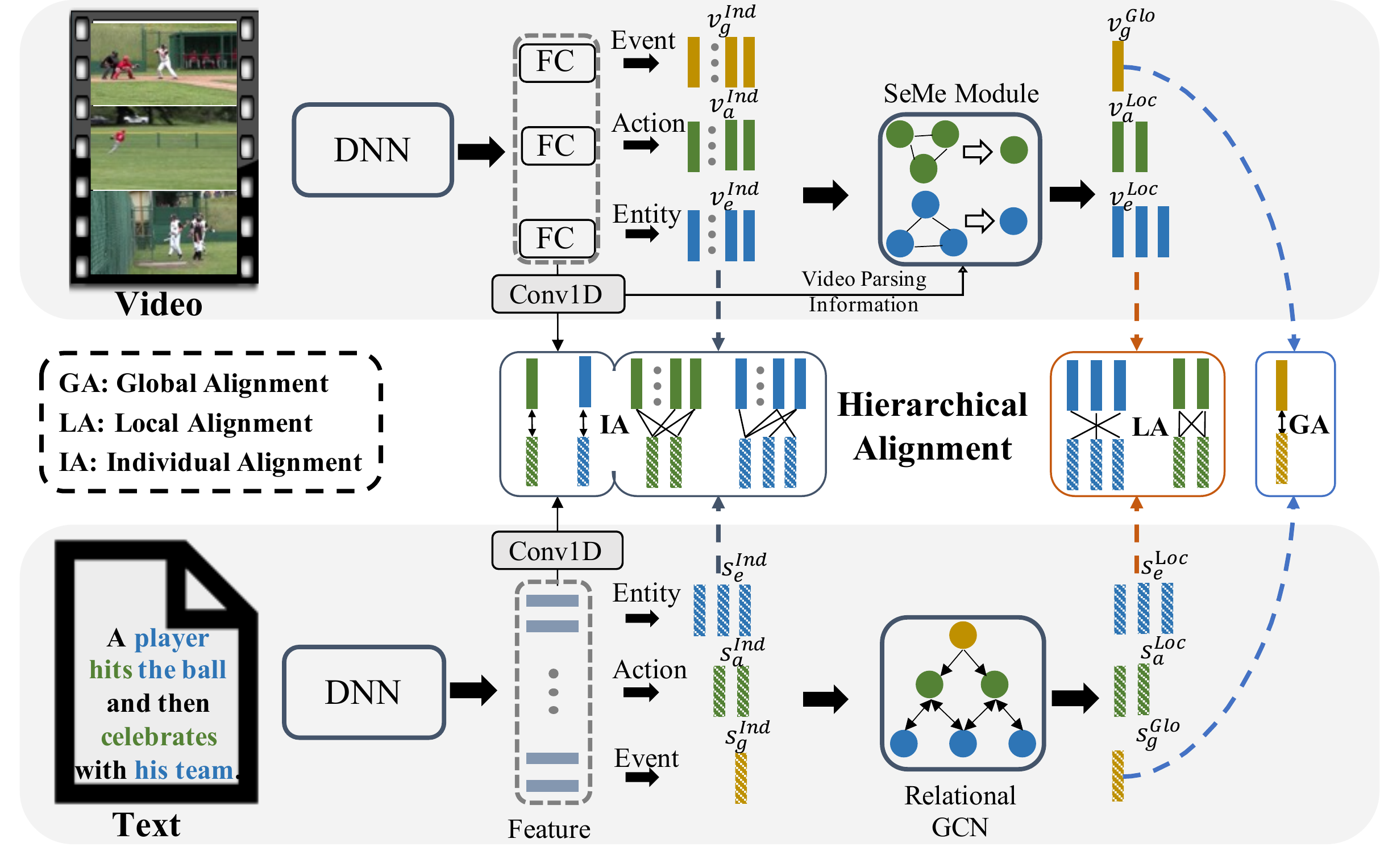}
  \caption{Framework of the proposed Hierarchical Alignment Network (HANet).}
  \label{framework}
\end{figure*}

\section{HANet}
We propose the hierarchical alignment network (HANet) for video-text retrieval, whose goal is to hierarchically align different level of video-text features and measure the similarity in different common spaces. The overview of HANet is illustrated in Figure \ref{framework}, which consists of four components: 1) Video-text parsing 
(Section \ref{sec:parsing}), i.e., parsing video and text with video-specific and text-specific parse manners respectively. 2) Video representations (Section
\ref{sec:videopre}), i.e., constructing frame, clip and video levels of features on the basis of video parsing.   3) Text representations
(Section \ref{sec:textpre}), i.e., constructing word, phrase, and sentence levels of features on the basis of text parsing. 4) Hierarchical alignment (Section \ref{sec:align}), i.e., aligning different semantic levels of video and text representations to compute their similarity. Finally, training and inference are introduced in Section \ref{sec:trainandinf}.

\subsection{Video-Text Parsing}
\label{sec:parsing}
Given a video $V$ of length $N$ and its corresponding caption $S$ of length $M$, we expect that not only global representations of the video and text are close in their common space, but also more fine-grained representations should be close. To achieve this goal, we first parse the video and text.

\subsubsection{Video Parsing} 

Unlike text parsing, video parsing is challenging since videos are more complicated but lacking the distinct semantic structure than texts. A possible way to video parsing is to introduce some existing vision operations, such as temporal segmentation, object detection, tracking, which are time-consuming and complex for practice application. To address this issue, we propose to project video frames into the concept space with the predefined concept vocabulary, where each frame is explicitly associated with the specific concepts. That is, we establish cross-modal associations between video frames (or clips) with 
key words (such as nouns and verbs) via concepts. 

Formally, given the predefined action concept vocabulary of size $K_a$, we project the corresponding action feature map $v^{Ind}_a\in \mathbb{R}^{N \times D^v}$ of video into the $K_a$ dimensional action concept space using CNN. Likewise, we adopt another CNN to project entity feature map $v^{Ind}_e\in \mathbb{R}^{N \times D^v}$ into the $K_e$ dimensional entity concept space. That is,
\begin{equation}
  l^v_a= \sigma\left(BN\left(Conv1d_{k=5}\left(v^{Ind}_a\right)\right)\right)
\end{equation}
\begin{equation}
  l^v_e= \sigma\left(BN\left(Conv1d\left(v^{Ind}_e\right)\right)\right)
\end{equation}
where $l^v_a\in \mathbb{R}^{N \times K_a}$ and $l^v_e \in \mathbb{R}^{N \times K_e}$ are denoted as the confidence for action and entity concepts, respectively. $\sigma$ is the sigmoid activation, $BN$ is the batch normalization. The detailed descriptions of $v^{Ind}_a$ and $v^{Ind}_e$ can be referred to Section \ref{sec:videopre}.
It is worth mentioning that we employ the convolution layer with kernel size of $k=5$ to obtain the probabilistic output of action concept, which is to capture the intrinsic motion information in continuous video frames.

\subsubsection{Text Parsing} 
Following the prior work \cite{chen2020fine}, we employ the off-the-shelf semantic role parsing toolkit \cite{shi2019simple} to obtain verbs, nouns as well as the semantic role of each noun to the corresponding verb. We refer the reader to \cite{shi2019simple, chen2020fine} for detailed descriptions. Here, verbs are considered as actions, likewise, nouns are entities.
Then, we project sentence words into the action and entity concepts, and their confidences are obtained as follows,
\begin{equation}
  l^s_a= \sigma\left(BN\left(Conv1d\left(s^{Ind}_a\right)\right)\right)
\end{equation}
\begin{equation}
  l^s_e= \sigma\left(BN\left(Conv1d\left(s^{Ind}_e\right)\right)\right)
\end{equation}
where $l^s_a\in \mathbb{R}^{M \times K_a}$ and $l^s_e \in \mathbb{R}^{M \times K_e}$ are denoted as the confidence for action and entity concepts, respectively. The detailed descriptions of $s^{Ind}_a$ and $s^{Ind}_e$ can be referred to Section \ref{sec:textpre}.

Video-text parsing introduces a new concept-based match space between videos and texts on action and entity concept-level. In this way, good interpretability is also introduced.

\subsection{Video Representations}

\label{sec:videopre}
We design three different granularities of representations, namely individual, local and global, corresponding to the video frame, video clip, and whole video respectively, which capture fine-to-coarse information and are complementary to each other.

\subsubsection{Individual-level Representation}
Formally, given the video $V$, we first use the pre-trained CNN to extract frame-level features $F^v=\{f^v_i\}^N$. Following \cite{chen2020fine}, we then employ different fully connected (FC) layers to encode the video into three semantic-level embeddings as follows:
\begin{equation}
  v^{Ind}_x= FC\left(F^v\right),\quad  x \in \{a,e, g\}
\end{equation}
where $v^{Ind}_x \in \mathbb{R}^{N \times D^v}$.
Since $v^{Ind}_a$ and $v^{Ind}_e$ only contain frame-level information, here, they are considered as individual-level representations. $v^{Ind}_g$ is used for the global-level representation in the later section.

\subsubsection{Local-level Representation} 
To further explore the contextual information between frames, we propose a Selecting and Merging (SeMe) module to generate the local-level representation. The SeMe module takes individual-level representation as input, and outputs concept confidence. To extract more high-level representation, we employ a convolution layer to project $v^{Ind}_a$ and $v^{Ind}_e$ into another spaces. That is,
\begin{equation}
  v^{SE}_a= SE\left(v^{Ind}_a\right)
\end{equation}
\begin{equation}
  v^{SE}_e= SE\left(v^{Ind}_e\right)
\end{equation}
where $SE$ is the simple yet effective Squeeze-and-Excitation block \cite{hu2018squeeze} to explicitly model inter-dependencies between channels. After that, we obtain the local-level representation with the help of video parsing. Specifically, each frame has $K_a$ dimensional action concept confidence and $K_e$ dimensional entity concept confidence, here we expect to know which action and entity concepts are associated with each video rather than each frame. To this end, we introduce a multiple instance learning (MIL) based mechanism inspired by \cite{paul2018w, wu2020not, wu2021learning}, which can be presented as follows,
\begin{equation}
  p^v_{a,i} = \frac{1}{\tau} \sum_{\forall z \in Z_i} z
  \label{mil}
\end{equation}
where $\tau=\lfloor \frac{N}{8} \rfloor$, and $Z_i$ is the set of $\tau$-max frame-level confidence scores for the $i^{th}$ action concept, which is selected from frame-level action concept confidence $l^v_a$, and the size of $Z_i$ is $\tau$. $p^v_a$ is the video-level action concept confidence given the video $V$. For the $i^{th}$ action concept, we obtain the $i^{th}$ video-level action concept confidence $p^v_{a,i}$ by averaging $Z_i$.
In the similar vein, we obtain $p^v_e$. Given $p^v_a$ and $p^v_e$, $N_a$ and $N_e$ action and entity concepts with the highest confidence are selected as the reliable concepts for the video $V$. For each selected action concept, we select the video clip of size $k=5$ in $v^{SE}_a$ which corresponds to the highest confidence, and employ the average pooling to obtain a feature vector. Through this operation, we finally obtain the local-level action representation $v^{Loc}_a \in \mathbb{R}^{N^a \times D^v}$. The main difference between obtaining $v^{Loc}_a$ and obtaining local-level entity representation $v^{Loc}_e$ is that we select 3 frames in $v^{SE}_e$, which may not be continuous, corresponding to the top 3 highest confidences and obtain a feature vector by the average pooling. By merging adjacent and semantically similar frames, local-level representations contain more rich information and capture local range dependencies, they can explicitly align with nouns and verbs in texts. 

To summarize, local-level representation is the aggregation of several frame-level features corresponding to reliable concepts. First, we obtain the video-level concept confidence based on frame-level concept confidence via Equation (\ref{mil}). Then, we select highly confident action and entity concepts as reliable concepts based on video-level concept confidence. For each reliable concept, we choose some highly confident frame features based on the frame-level concept confidence, and aggregate them to generate the final local-level representation.

\subsubsection{Global-level Representation} 
For the global event level, we adopt an attention mechanism to average the frame-level features as a single global vector $v^{Glo}_g$ that represents the salient event in the video, which is presented as follows,
\begin{equation}
  v^{Glo}_g = \alpha^{\top}v^{Ind}_g 
  \label{att0}
\end{equation}
\begin{equation}
  \alpha = softmax\left(v^{Ind}_gW\right)
  \label{att1}
\end{equation}
where $\alpha$ is the attention score, and $W$ is the learnable weight.

\subsection{Text Representations}
\label{sec:textpre}
Parallel to video representations, text representations consist of three levels of granularity, i.e., the individual level corresponds to the word, the local level corresponds to the context, and the global level corresponds to the sentence. Three levels of text representations are used to align with corresponding video representations in the hierarchical alignment. Formally, given the sentence $S$, we employ the pre-trained model to extract the word embeddings $F^s=\{f^s_i\}^M$, then generate three different levels of representations on top of text parsing.

\subsubsection{Individual-level Representation.} We utilize an bidirectional GRU (Bi-GRU) \cite{cho2014learning} to generate a sequence of contextual-aware word embeddings as follows,
\begin{equation}
  \overrightarrow{s^{Ind}}= \overrightarrow{GRU}\left(F^s\right)
\end{equation}
\begin{equation}
  \overleftarrow{s^{Ind}}= \overleftarrow{GRU}\left(F^s\right)
\end{equation}
\begin{equation}
  s^{Ind} = \left(\overrightarrow{s^{Ind}}+\overleftarrow{s^{Ind}}\right)/2
\end{equation}

Based on the text parsing, we select features corresponding to verbs and nouns as the individual-level representations $s^{Ind}_a \in \mathbb{R}^{M^a \times D^s}$ and $s^{Ind}_e \in \mathbb{R}^{M^e \times D^s}$ that are subsets of $s^{Ind}$, which  correspond to $v^{Ind}_a$ and $v^{Ind}_e$ in videos. Meanwhile, we employ the attention mechanism similar to Equations (\ref{att0})-(\ref{att1}) to obtain the global event embedding $s^{Ind}_g$ for the following local and global-level representations.

\subsubsection{Local and Global-level Representation.} We follow the pioneer work \cite{chen2020fine} and employ modified relational GCN \cite{schlichtkrull2018modeling} to obtain local and global-level representations. To be specific, we gather three features, i.e., $s^{Ind}_g$, $s^{Ind}_a$ and $s^{Ind}_e$, and use $g \in \mathbb{R}^{\left(1+M^a+M^e\right) \times D^s}$ to denote this feature set, which is initialized node embeddings of graph. Different semantic roles by text parsing are the edges of graph.

Here, we only use one GCN layer, which is presented as follows,
\begin{equation}
  g^{1}_i = g^{0}_i+\sum_{j \in N_i}\left(\beta_{ij}\left(W_t \odot W_rr_{ij} \right)g_j\right)
\end{equation}
where $W_t \in \mathbb{R}^{D^s \times D^s}$ is the transformation matrix, $W_r \in \mathbb{R}^{D^s \times K^r}$ is role embedding matrix, $K^r$ is the number of semantic roles, $N_i$ is neighborhood nodes of node $i$, $r_{ij}$ an one-hot vector of length $K^r$ denoting the edge type from node $i$ to $j$, and $\beta_{ij}$ is the similarity between node $i$ to $j$, which is computed as follows,
\begin{equation}
  \beta_{ij} = softmax\left(\frac{\phi\left(g_i\right)\varphi\left(g_j\right)^{\top}}{\sqrt{D^s}}\right)
\end{equation}
here $\phi\left(g_i\right)=W_{\phi}g_i$ and $\varphi\left(g_j\right)=W_{\varphi}g_j$ are two embeddings. We refer the reader to \cite{chen2020fine} for detailed descriptions of the modified relational GCN.

The outputs from the GCN layer are the final different levels of representations, which are denoted as $s^{Loc}_a$ for local-level action representation, $s^{Loc}_e$ for local-level entity representation and $s^{Glo}_g$ for global-level representation. Notably, we do not use the relational GCN in videos since the semantic role of each entity to the corresponding action is unknown.

\subsection{Hierarchical Alignment}
\label{sec:align}
After the aforementioned text encoding and video encoding, we obtain three levels of representations, namely, individual, local, and global levels. In this section, we introduce how to hierarchically align representations at three different level.

\subsubsection{Individual Alignment.} Since there are multiple components in the video and text at the individual level, following \cite{lee2018stacked, chen2020fine, diao2021similarity}, we use the stack attention mechanism to align multiple components and compute the overall similarity score. For the sake of clarity, we use $v^{Ind}$ to denote $v^{Ind}_a$ and $v^{Ind}_e$, in the similar vein, we define $s^{Ind}$. We use cosine similarity to compute similarities between each pair of cross-modal components  $c^{Ind}_{ij} = cos\left(v^{Ind}_i, s^{Ind}_j\right)$. Then we compute the attention weight that dynamically aligns sentence words and video frames as follows,
\begin{equation}
  \gamma_{ij}= softmax\left(\lambda \lbrack c^{Ind}_{ij}\rbrack_+ / \sqrt{\sum_{j} \lbrack c^{Ind}_{ij}\rbrack^2_+} \right)
  \label{stack0}
\end{equation}
where $\lambda$ is the temperature parameter, $\lbrack \cdot \rbrack_+ \equiv max(\cdot,0)$. The final similarity summarizes all individual component similarities and is shown as follows,
 \begin{equation}
  c^{Ind} = \sum_i\left(\sum_j{\gamma_{ij}c^{Ind}_{ij}}\right)
  \label{stack1}
\end{equation}

More importantly, we introduce the concept-based similarity based on the concept confidence, where the concept confidence develops from individual-level representations. For simplicity, we use $p^v$ to denote $p^v_a$ and $p^v_e$. Following \cite{dong2021dual}, we employ generalized Jaccard similarity to compute the concept-based similarity,
\begin{equation}
  c^p = \frac{\sum_i min\left(p^v_i, p^s_i\right)}{\sum_i max\left(p^v_i, p^s_i\right)}
\end{equation}
where $p^s$ is denoted $p^s_a$ and $p^s_e$, which is obtained as in Equation (\ref{mil}).

\subsubsection{Local Alignment.} For the local level, we employ the stack attention mechanism similar to Equations (\ref{stack0})-(\ref{stack1}) to obtain $c^{Loc}_a$ and $c^{Loc}_e$. 

\subsubsection{Global Alignment.} At the global event level, the video
and text are encoded into global vectors. We use the cosine similarity to measure the cross-modal similarity between global video and global text $c^{Glo}_{g} = cos\left(v^{Glo}_g,s^{Glo}_g\right)$.

\subsection{Training and Inference.}
\label{sec:trainandinf}
\subsubsection{Training.} Once all similarity scores are computed, we obtain the two similarity between the video $V$ and the sentence $S$, that is,
\begin{equation}
  c_l\left(V,S\right) = \left(c^{Ind}_a+c^{Ind}_e+c^{Loc}_a+c^{Loc}_e+c^{Glo}_g\right)/5
\end{equation}
\begin{equation}
  c_p\left(V,S\right) = \left(c^p_a+c^p_e\right)/2
\end{equation}

The widely used ranking loss with hard negative sampling strategy is used to optimize HANet, here $\mathcal{L}_l$ and $\mathcal{L}_p$ are presented as follows,
\begin{equation}
  \mathcal{L}_l = \lbrack \Delta+c_l\left(V, S^-\right)-c_l\left(V, S\right)\rbrack_++\lbrack \Delta+c_l\left(V^-, S\right)-c_l\left(V, S\right)\rbrack_+
\end{equation}
\begin{equation}
  \mathcal{L}_p = \lbrack \Delta+c_p\left(V, S^-\right)-c_p\left(V, S\right)\rbrack_++\lbrack \Delta+c_p\left(V^-, S\right)-c_p\left(V, S\right)\rbrack_+
\end{equation}
where ($V$, $S$) are the positive pair, and $V^-$ and $S^-$ are the hardest negatives in a mini-batch.

Besides, we use the binary cross-entropy (BCE) loss for concept learning,
\begin{equation}
\begin{split}
  \mathcal{L}_a = -\frac{1}{K^a} \sum_{i}\left(y_{a,i} log\left(p^v_a\right) + (1-y_{a,i}) log\left(1-p^v_a\right)\right)\\
  - \frac{1}{K^a} \sum_{i}\left(y_{a,i} log\left(p^s_a\right) + (1-y_{a,i}) log\left(1-p^s_a\right)\right)
\end{split}
\end{equation}

\begin{equation}
\begin{split}
  \mathcal{L}_e = -\frac{1}{K^e} \sum_{i}\left(y_{e,i} log\left(p^v_e\right) + (1-y_{e,i}) log\left(1-p^v_e\right)\right)\\
  - \frac{1}{K^e} \sum_{i}\left(y_{e,i} log\left(p^s_e\right) + (1-y_{e,i}) log\left(1-p^s_e\right)\right)
\end{split}
\end{equation}
where $y_{a}$ and $y_{e}$ are the ground-truth.

The overall loss is the combination of aforementioned losses, which is shown as follows,
\begin{equation}
  \mathcal{L}_{total} = \mathcal{L}_l + \eta\mathcal{L}_p + \mu\left(\mathcal{L}_a + \mathcal{L}_e\right)
\end{equation}
where $\eta$ and $\mu$ is the trade-off hyper-parameters.

\subsubsection{Inference.} we simply take the average of $c_p\left(V,S\right)$ and $c_l\left(V,S\right)$ as the final video-text similarity between $V$ and $S$ for video-text retrieval.

\begin{table*}
  \caption{Comparisons with the state-of-the-art methods on the MSR-VTT dataset.}
  \label{tab:msrvtt}
  \begin{tabular}{l|cccc|cccc|c}
    \toprule
    \multirow{2}{*}{Method}  & \multicolumn{4}{c|}{Text-to-Video} & \multicolumn{4}{c|}{Video-to-Text} & \multirow{2}{*}{SumR}\\
    ~ & R@1 & R@5 & R@10 & MdR & R@1 & R@5 & R@10 & MdR \\
    \hline
    VSE \cite{kiros2014unifying} & 5.0 & 16.4 & 24.6 &47 & 7.7 & 20.3 & 31.2 & 28 & 105.2 \\
    VSE++ \cite{faghri2017vse++} & 5.7 & 17.1 & 24.8 & 65 & 10.2 & 25.4 & 35.1 & 25 & 118.3 \\
    Mithum et al. \cite{mithun2018learning} & 5.8 & 17.6 & 25.2 & 61 & 10.5 & 26.7 & 35.9 & 25 & 121.7 \\
    W2VV \cite{dong2018predicting}& 6.1 & 18.7 & 27.5 & 45 & 11.8 & 28.9 & 39.1 & 21 & 132.1 \\
    Dual Encoding \cite{dong2019dual} & 7.7 & 22.0 & 31.8 & 32 & 13.0 & 30.8 & 43.3 & 15 & 148.6 \\
    TCE \cite{yang2020tree} &7.7 & 22.5 & 32.1 & 30 & - & - & - & - & - \\
    Zhao et al.\cite{zhao2020stacked} & 8.8 & 25.5 & 36.5 & 22 & 14.0 & 33.1 & 44.9 & 14 & 162.8 \\
    HGR \cite{chen2020fine} & 9.2 & 26.2 & 36.5 & 24 & 15.0 & 36.7 & 48.8 & 11 & 172.4 \\
    \hline
    \textbf{HANet} & \textbf{9.3} & \textbf{27.0} & \textbf{38.1} & \textbf{20} & \textbf{16.1} & \textbf{39.2} & \textbf{52.1}	& \textbf{9} & \textbf{181.8} \\
    \bottomrule
  \end{tabular}
\end{table*}

\begin{table*}
  \caption{Comparisons with the state-of-the-art methods on the VATEX dataset.}
  \label{tab:vatex}
  \begin{tabular}{l|cccc|cccc|c}
    \toprule
    \multirow{2}{*}{Method}  & \multicolumn{4}{c|}{Text-to-Video} & \multicolumn{4}{c|}{Video-to-Text} & \multirow{2}{*}{SumR}\\
    ~ & R@1 & R@5 & R@10 & MdR & R@1 & R@5 & R@10 & MdR \\
    \hline
    W2VV \cite{dong2018predicting} & 14.6 & 36.3 & 46.1 & - & 39.6 & 69.5 & 79.4 & - & 285.5 \\
    VSE++ \cite{faghri2017vse++} & 31.3 & 65.8 & 76.4 & - & 42.9 & 73.9 & 83.6 & - & 373.9 \\
    CE \cite{liu2019use} & 31.1 & 68.7 & 80.2 & - & 41.3 & 71.0 & 82.3 & - & 374.6 \\
    W2VV++ \cite{li2019w2vv++} & 32.0 & 68.2 & 78.8 & - & 41.8 & 75.1 & 84.3 & - & 380.2 \\
    Dual Encoding \cite{dong2019dual} & 31.1 & 67.4 & 78.9 & - & - & - & - & - & - \\
    HGR \cite{chen2020fine} & 35.1 & 73.5 & 83.5 & - & - & - & - & - & - \\
    HSL \cite{dong2021dual} & \textbf{36.8} & 73.6 & 83.7 & - & 46.8 & 75.7 & 85.1 & - & 401.7 \\
    \hline
    \textbf{HANet} & 36.4 & \textbf{74.1} & \textbf{84.1} & \textbf{2} & \textbf{49.1} & \textbf{79.5} & \textbf{86.2} & \textbf{2} & \textbf{409.4} \\
    \bottomrule
  \end{tabular}
\end{table*}

\section{Experiments}

In this section, we first introduce two datasets (MSR-VTT \cite{xu2016msr} and VATEX \cite{wang2019vatex}) performed in our experiments. Then we compare our HANet with recent state-of-the-art methods and analyze its effectiveness. We also investigate each component in our HANet by ablation studies.

\subsection{Datasets and Evaluation Metrics}

\subsubsection{Datasets}
\textbf{MSR-VTT} dataset \cite{xu2016msr} is composed of 10000 video clips with 20 text descriptions per clip. We follow the official data split, where 6573, 497 and 2990 videos are used for training, validation and testing, respectively. \textbf{VATEX} dataset \cite{wang2019vatex} is a large-scale bilingual video description dataset, each clip is accompanied by 10 English text descriptions and 10 Chinese text descriptions. Here, 
we only use the English text descriptions. Following the partition provided by \cite{chen2020fine, dong2021dual}, we use 25991 video clips for training, 1500 clips for validation and 1500 clips for testing, where validation and test sets are obtained from the official validation set since the annotations on test set are private.

\subsubsection{Evaluation Metrics} 
Following prior works, we report the results using the rank-based performance metric, i.e., Recall at K (R@K, K=1, 5, 10, higher is better), Median Rank (MdR, lower is better), and Sum of all Recalls (SumR, higher is better) to measure the overall performance.

\subsection{Implementation Details}

\subsubsection{Video-Text Features}
For fair comparison, we apply the same feature in our HANet to all the compared methods. For MSR-VTT, we utilize the visual feature provided by \cite{chen2020fine} with dimension of 2048, which is extracted with ResNet152 pre-trained on ImageNet \cite{he2016deep}. For VATEX, we use the officially provided I3D \cite{carreira2017quo} video feature. For the text features on both MSR-VTT and VATEX, we set the word embedding size as 300 and initialize with pre-trained Glove embeddings \cite{pennington2014glove}.

\subsubsection{Concept Vocabulary} The concept vocabulary is constructed from all training sentences. Specifically, we first remove all English stop-words and punctuations, and use NLTK toolkit \footnote{http://www.nltk.org} to find the part-of-speech tags. After that, only nouns and verbs are retained, which correspond to the entities and actions, respectively. To avoid duplication of concepts, we also lemmatize these nouns and verbs by NLTK. Finally, the top $K_a=$512 frequent verbs and $K_e=$1024 nouns are selected as the final action and entity concept vocabularies, respectively. Based on data statistics of training samples, we found that the top $K_a=$512 frequent verbs and $K_e=$1024 nouns cover the vast majority of high-frequency words.

\subsubsection{Training} We implement HANet using PyTorch \footnote{https://pytorch.org} on the NVIDIA V100 GPU. We use Adam \cite{kingma2014adam} to optimize HANet, with learning rate of 1e-4 and batch size of 64. The maximal number of epochs is set to 50, and early stop occurs if the validation performance (SumR) does not improve in ten consecutive epochs. As for hyper-parameters, $N_a$ and $N_e$ are set to 10 and 20 respectively based on data statistics of training samples. The weight $\eta$ and $\mu$ in the combined losses is empirically set to 0.1 and 0.01, respectively. Following \cite{chen2020fine}, the temperature parameter $\lambda$ is set to 4, and the margin $\Delta$ is set to 0.2.

\begin{table*}
  \caption{Ablation studies on the MSR-VTT dataset to investigate the effectiveness of hierarchical alignment.}
  \label{tab:ha}
  \begin{tabular}{ccc|cccc|cccc|c}
    \toprule
    Individual  & Local  & Global &\multicolumn{4}{c|}{Text-to-Video} & \multicolumn{4}{c|}{Video-to-Text} & \multirow{2}{*}{SumR}\\
    Alignment & Alignment & Alignment & R@1 & R@5 & R@10 & MdR & R@1 & R@5 & R@10 & MdR \\
    \hline
     $\surd$ &  &  & 8.6 & 25.7 & 36.4 & 23 &14.1 & 36.4 & 48.6 & 11 & 169.8 \\
     & $\surd$ &  & 8.4 & 24.6 & 34.9 & 25 &13.2 & 35.2 & 47.0 & 12 & 163.3 \\
      &  & $\surd$ & 8.4 & 24.9 & 35.6 & 24 &14.3 & 33.9 & 46.1 & 13 & 163.2 \\
    \hline
    $\surd$ & $\surd$ &  & 9.2 & 26.7 & 37.6 & 21 &15.6 & 38.8 & 51.4 & 10 & 179.3 \\
    $\surd$ &  & $\surd$ & 9.1 & 26.6 & 37.8 & 21 &15.5 & 38.0 & 50.6 &10 & 177.6 \\
     & $\surd$ & $\surd$ & 8.6 & 25.3 & 36.2 & 23 & 13.6 & 35.6 & 48.0 &11 & 167.3 \\
    \hline
    $\surd$ & $\surd$ & $\surd$ & \textbf{9.3} & \textbf{27.0} & \textbf{38.1} & \textbf{20} & \textbf{16.1} & \textbf{39.2} & \textbf{52.1}	& \textbf{9} & \textbf{181.8} \\
    \bottomrule
  \end{tabular}
\end{table*}

\begin{table*}
  \caption{Ablation studies on the MSR-VTT dataset to investigate the effectiveness of SeMe and relational GCN.}
  \label{tab:ab}
  \begin{tabular}{l|cccc|cccc|c}
    \toprule
    \multirow{2}{*}{Model}  & \multicolumn{4}{c|}{Text-to-Video} & \multicolumn{4}{c|}{Video-to-Text} & \multirow{2}{*}{SumR}\\
    ~ & R@1 & R@5 & R@10 & MdR & R@1 & R@5 & R@10 & MdR \\
    \hline
    FC $\to$ SeMe & 9.2 & 26.7 & 37.8 & 21 & 14.7 & 38.3 & 50.9 & 10 & 177.6 \\
    FC $\to$ relational GCN & 8.9 & 26.6 & 37.6 & 21 & 14.9 & 38.1 & 50.7 & 10 & 176.8 \\

    \textbf{HANet} & \textbf{9.3} & \textbf{27.0} & \textbf{38.1} & \textbf{20} & \textbf{16.1} & \textbf{39.2} & \textbf{52.1}	& \textbf{9} & \textbf{181.8} \\
    \bottomrule
  \end{tabular}
\end{table*}

\subsection{Comparison with State-of-the-Art Methods}
We compare our HANet with state-of-the-art methods on the MSR-VTT and VATEX datasets. Notably, all comparison methods take the same visual feature as input. The comparison results on the MSR-VTT are shown in Table~\ref{tab:msrvtt}. We observe that HANet achieves significant performance improvements over comparison methods on text-to-video and video-to-text retrieval tasks. All baselines except HGR \cite{chen2020fine} only use the global features to compute the similarity between the video and text. HGR is similar to our HANet among these baselines, which also decomposes the text and aligins the video and text in a global-to-local fashion. Compared with HGR, our HANet is equipped with concept-guiding video parsing, and aligns videos and texts in a more fine-grained and more precise fashion, i.e., individual level (frame and word), local level (clip and phrase), and global level (video and sentence). As a consequence, HANet significantly outperforms HGR in all evaluation metrics, especially, boosts the overall retrieval quality by a margin of 9.4 in SumR.

To demonstrate the reliability of HANet, we also carry out experiments on another dataset, i.e., VATEX, using different visual features I3D. From results in Table~\ref{tab:vatex}, it is also easy to notice that HANet outperforms other existing methods by a large margin. Here HSL \cite{dong2021dual} stresses the importance of video-text representation and also uses different semantic levels of features. However, HSL simply concatenates all levels of features and overlooks the hierarchical alignment.

\begin{figure*}[htb]
  \centering
  \includegraphics[width=15cm]{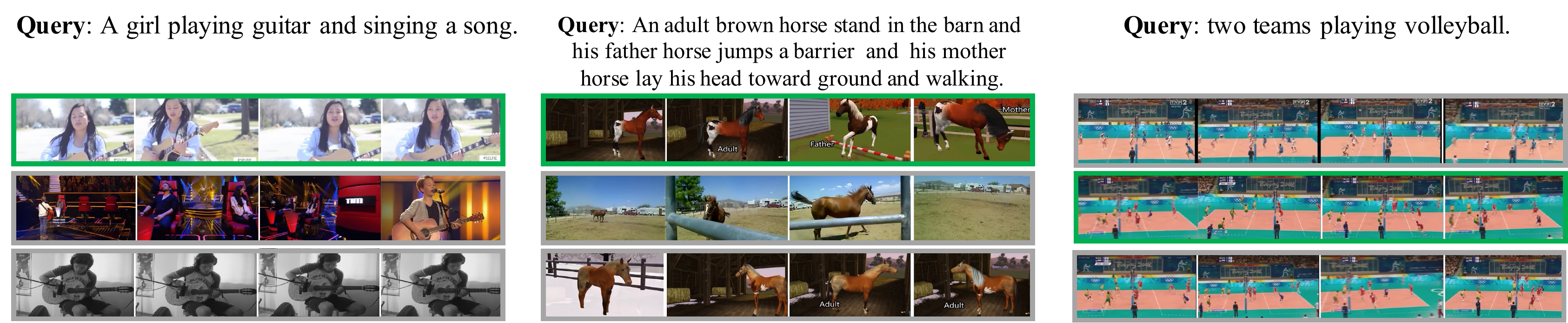}
  \caption{Top 3 text-to-video retrieval examples on the MSR-VTT test set. (green box: correct; gray box: incorrect)}
  \label{qualitative}
\end{figure*}

\begin{figure*}[htb]
  \centering
  \includegraphics[width=13cm]{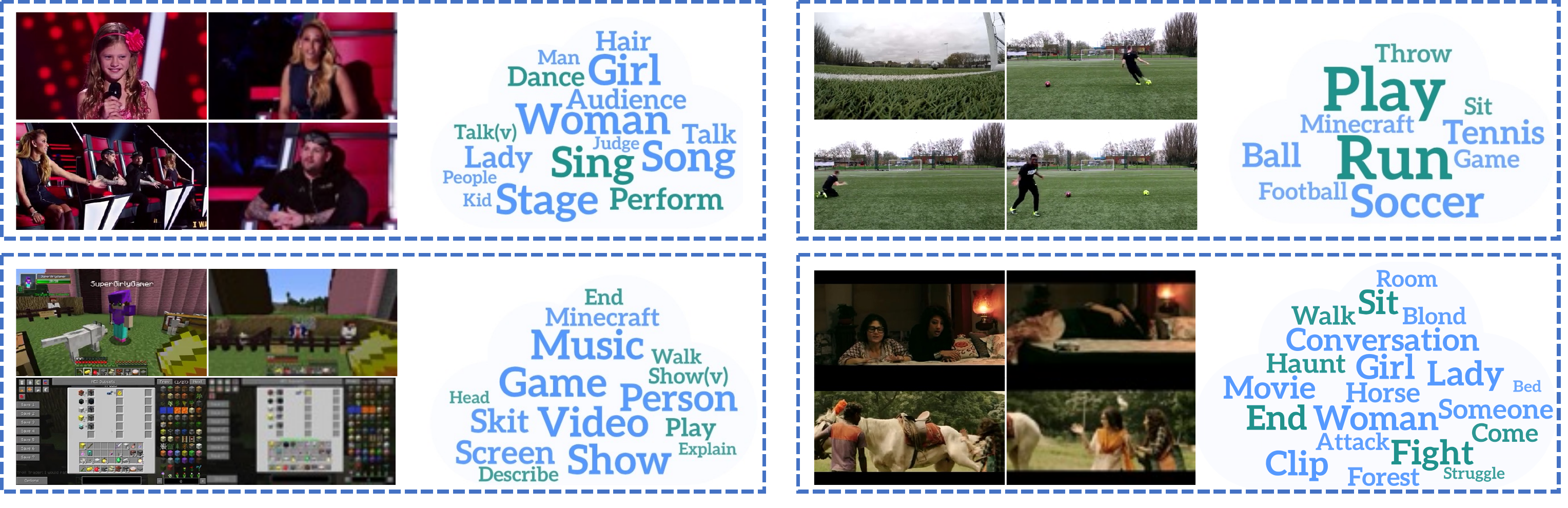}
  \caption{Examples of concept prediction in video parsing. Four different videos are taken from the MSR-VTT test set. Bigger font meaning larger predicted scores in each sample. (green: action concepts; blue: entity concepts)}
  \label{concept}
\end{figure*}

\begin{figure*}[htb]
  \centering
  \includegraphics[width=13cm]{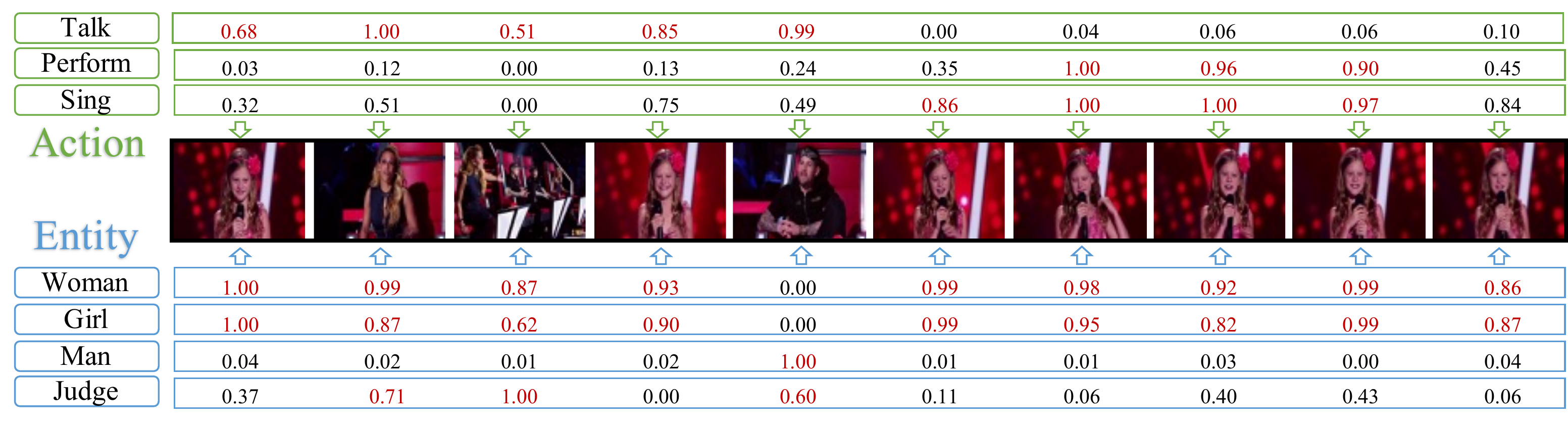}
  \caption{Examples of  concept prediction of video frames in video parsing. A video is taken from the MSR-VTT test set. We normalize the concept confidence to [0,1] for each concept.}
  \label{concept_assign}
\end{figure*}

\subsection{Ablation Studies}
In this section, we conduct experiments on the MSR-VTT to verify the effectiveness of each component in HANet.

\subsubsection{Effectiveness of Hierarchical Alignment}
We first investigate the effectiveness of hierarchical alignment. The results of using different alignments of our HANet are shown in Table~\ref{tab:ha}. As we can see that, only using a single alignment achieves worse performance. In particular, only using global alignment is similar to most previous works that simply use a single representation, which results in an obvious drop of 18.6 in terms of SumR. It convincingly demonstrates the improvement contributed by the proposed hierarchical alignment which provides more fine-grained information. Besides, we notice that any combination of two different alignments can gain performance improvements, which demonstrates all three alignments are effective. Finally, the combination of all three alignments further improves performance, which demonstrates individual, local and global-level information are complementary.

\subsubsection{Effectiveness of Local-level Representation}
Here we use two FC layers to replace our designed SeMe module in terms of video and relational GCN in terms of text, respectively. We show results in Table~\ref{tab:ab}. Compared with HANet, they both suffer performance degradation, which convincingly demonstrates that, 1) video contextual information is introduced by selecting and merging concept-based relevant frames in our SeMe module; 2) text contextual information is also captured by relational GCN.

\subsection{Qualitative Analyses}
\subsubsection{Visualization of the Text-to-Video Retrieval}
We visualize several examples on the MSR-VTT test set for text-to-video retrieval in Figure \ref{qualitative}. In the left and middle samples, our HANet successfully retrieve the correct video given query. The right sample shows an ambiguous case, where all top 3 retrieved videos present a scene of "two teams playing volleyball". We argue that this was caused by the instance-based assumption \cite{wray2021semantic} in current video-text retrieval, namely only a single video is relevant to a query. In fact, these three videos can be deemed equally relevant.

\subsubsection{Visualization of Action- and Entity-level Concept} Since video parsing in HANet is based on the concept-based classification, in this part, we visualize some examples of concept prediction at the video and frame-level. In Figure \ref{concept}, we observe that some relevant concepts are predicted with high confidence, for example, "play", "soccer" in the example of the top right corner. However, there are also some irrelevant predicted concepts, such as "tennis" and "throw". In general, concepts predicted by HANet are reasonable, and helpful for understanding the cross-modal retrieval.

The frame-level concept prediction is shown in Figure \ref{concept_assign}, we highlight the areas with higher concept confidence in red. For the action concept, since we take as input the 5-frame clip to obtain the confidence of action concept, actions, e.g., "talk", "sing", contained in consecutive frames are retrieved. Intriguingly, based on the frame-level concept confidence, not only "woman", "girl" and "man" are retrieved, but also the "judge" that is gender-neutral is retrieved. which demonstrates the reliability of MIL based mechanism in Equation (\ref{mil}) for weakly supervised concept classification as well as the practicability of SeMe module.

\section{Conclusion}

In this paper, we propose the hierarchical alignment network (HANet) to make full use of complementary information of different semantic levels of representations for video-text retrieval. To this end, we first parse the video and text by concept-based weakly supervised classification and existing text parsing toolkit, respectively. Then we introduce the hierarchical alignment to align representations at the individual, local and global levels for computing cross-modal similarity. The quantitative and qualitative results on two popular text-video retrieval benchmarks significantly demonstrate the effectiveness of HANet. In the future, the more precise and efficient hierarchical alignment is yet to be explored since pairwise matching is relatively expensive.

\begin{acks}
This work is supported by the Key Project of Science and Technology Innovation 2030 supported by the Ministry of Science and Technology of China under Grant 2018AAA0101302 and the General Program of National Natural Science Foundation of China (NSFC) under Grant 61773300 and by Alibaba Group through Alibaba Research Intern Program.

\end{acks}

\bibliographystyle{ACM-Reference-Format}
\bibliography{sample-base}

\end{document}